\documentclass[10pt,a4paper]{article}

\usepackage{booktabs} % For formal tables
\usepackage{subcaption,graphicx}
\usepackage{appendix,url,authblk}
\usepackage{amsmath,amssymb}
\title{Large-scale image analysis using docker sandboxing}

\author[1,2]{B Sengupta\thanks{bs573@cam.ac.uk; BS has a dual appointment at Dept. of Bioengineering, Imperial College London.}}
\author[1]{E Vazquez}
\author[1]{M Sasdelli}
\author[1]{Y Qian}
\author[1]{M Peniak}
\author[1]{L Netherton}
\author[1]{G Delfino}
\affil[1]{Cortexica Vision Systems Limited, London SE1 8RT, UK}
\affil[2]{Dept. of Engineering, University of Cambridge, Cambridge CB2 1PZ, UK}

\date{}
\begin{document}

\maketitle

\begin{abstract} 
With the advent of specialized hardware such as Graphics Processing Units (GPUs), large scale image localization, classification and retrieval  have seen increased prevalence. Designing scalable software architecture that co-evolves with such specialized hardware is a challenge in the commercial setting. In this paper, we describe one such architecture (\textit{Cortexica}) that  leverages scalability of GPUs and sandboxing offered by docker containers. This allows for the flexibility of mixing different computer architectures as well as computational algorithms with  the security of a trusted environment. We illustrate the utility of this framework in a commercial setting i.e., searching for multiple products in an image by combining image localisation and retrieval. 
\end{abstract}

\section{Introduction}

Large-scale image retrieval has been a mainstay for both academic research and commercial products for several years \cite{la1998combining, perd2009efficient, perronnin2010large}. A particular emphasis has been put for architectures that are scalable, redundant and most importantly quicker than its predecessors  \cite{Zaharia2010,Gonzales2014,Esser2016}. This has resulted not only in  efficient methods for indexing or hashing large databases \cite{kulis2009kernelized, wang2012semi} but also automatic feature extraction using deep neural networks \cite{deng2009imagenet}. Tasks that relied on Viola-Jones \cite{viola2001rapid} or SVM based classifiers to produce  hand-crafted descriptors \cite{lin2011large, yang2009linear} have been replaced by deep learning approaches since the ImageNet 2012 competition \cite{russakovsky2015imagenet}. The emphasis has lately shifted from ``hand-crafted" feature descriptors to ``hand-crafted" network architectures. Unlike statistical learning theory \cite{Vapnik1998} only a handful of studies have attempted to move on from this burgeoning resurgence of ``deep-learning" architectures to  understanding the basic principles on which such networks are based \cite{Mallat2016,Dauphin2014,Montufar2014}. Nevertheless,  deep-neural networks have had immense commercial success.

Whereas efficient algorithms are unequivocally essential to achieve efficient classification, localisation and retrieval, commercial applications introduce new challenges. For commercial applications, a combination of efficient algorithms with a flexible and scaleable architecture is necessary. This paper describes one such architecture (\textit{Cortexica})  for large-scale image localisation and retrieval in a cost-effective and time-efficient manner. Our aim is to present the architecture and tools needed to develop a fast, scalable and flexible  system by focusing on a particular commercial application -- a multi-product search algorithm. This application is explained in Section \ref{sec:localisation}. The architecture described has proven to be adequate on a commercial environment (\url{https://www.cortexica.com/}). An important property of our system is that is not limited to particular algorithms detailed in this paper. Thus, the same architecture can be used with different algorithms for classification, localisation or retrieval.

The rest of the paper is organised as follows. Our approaches for localisation and retrieval are detailed in Sections \ref{sec:localisation} and \ref{sec:retrieval}, respectively. Subsequently, in Section \ref{sec:Combining} we explain the combined framework for localisation and retrieval to finally draw conclusions in Section  \ref{sec:conclusions}.

\section{Multi-product search}
\label{sec:localisation}
It is typical in computer vision that a single image can have more than one item that needs to be identified; an additional aim can be to recommend a similar item from a vendor's database. Our approach is tailored for fashion items (clothes, accessories, etc.) yet the same approach can be followed for any type of objects, be it in medical imaging, transportation analytics, amongst many others. For example, we have  deployed our framework to analyze streaming video -- from walk-through at a fashion show. Our algorithm  is able to identify all of the clothing items that a  `fashion model' wears; additionally, returning similar items from a vendor's database for each identified item.

One methodology to achieve such an aim is to first localise the items of interest, i.e.
detect the items and find the associated bounding boxes, and then perform retrieval for each detected item. In the following section, we provide an overview of the different methods to perform object localisation as well as details on the architecture and the implementation for achieving good classification accuracy. 

The first step of the multi-product search is object localisation. Deep learning has proven to yield high accuracy at relatively low computational cost when using GPU-based computing \cite{Schmidhuber2015}. In this section, three networks for object localisation are introduced, and compared. The most relevant property of our deployment is that it allows us to hot-swap the neural network used (in a production environment), making it straight forward to introduce neural network modifications or for that matter other non-neural network algorithms.

\subsection{Image localisation approach}
\label{subsec:localisation_approach}

The state-of-the-art neural networks for object detection that we describe below include: Faster-R-CNN (Faster Region-based Convolutional Neural Network) \cite{ren15}, SSD (Single Shot Multi-Box Detector) \cite{liu15SSD}, and R-FCN (Region-based Fully Convolutional Networks) \cite{dai16rfcn}: \\

\textbf{Faster-RCNN}, a quasi-realtime object detection framework, is presented in \cite{ren15}.
The predecessor Fast RCNN \cite{girshick15fastrcnn} illustrated that  a convolution neural network (CNN) framework can be successfully applied for object detection, significantly increasing the probability of detecting an object. As a first step, the Visual Geometry Group (VGG) network is trained for image classification \cite{simonyanZ14a}, this is then further fine-tuned for object detection.
Faster R-CNN includes a region proposal network (RPN), that  builds upon Fast R-CNN  \cite{uijlings2013selective} enabling generation of fast proposals with little overhead.
The features of the convolutional part of VGG are shared between two tasks -- region proposal and classification. Just like the Spatial Pyramid Pooling (SPP) network \cite{he2014spatial}, the convolutional part of the network works just with images that have fixed aspect ratios. In our experience, based on  prior suggestions \cite{shrivastavaGG16}, training the classifier on a large number of `hard' negatives strengthens the classification performance and leads to better precision.\\

\textbf{Single Shot Multi-Box Detector (SSD)} \cite{liu15SSD}, uses an architecture that is equivalent to many class specific RPNs, each working on different feature maps. This to done to improve the detection of objects at different scales, thereby illustrating the benefits of a multi-scale architecture.
SSD uses a fully convolutional layer and an optimized VGG architecture wherein the input aspect ratios are fixed at 500x500 (similar to the classical VGG network). This allows for a competitive execution time, fast enough to be real-time, without compromising on the accuracy of Faster R-CNN. The primary disadvantage of this framework lies in the detection of small objects.\\

The \textbf{Region-based Fully Convolutional Network} \cite{dai16rfcn} has an architecture similar to Faster-RCNN, albeit without the fully connected layer of the network. Not having the fully connected layer allows us to calculate in a single forward pass the loss function for considerably large number of region proposals. These regions are then quickly sorted according to their loss function values, retaining only those regions that supersede a fixed threshold. These  are then utilized for gradient computation. This approach is called \textit{online hard example mining (OHEM)} \cite{shrivastavaGG16}; such a scheme is generally computationally infeasible to apply on a Faster-RCNN based network. In Faster-RCNN, for each region of interest (ROI), the network calculates a forward pass only when the images propagate to the fully connected layer. Therefore, if there are numerous ROIs, say more than a few hundred, it becomes increasingly time-consuming to train the network.

\begin{table*}
 \begin{center}
  \begin{tabular}{ l | l l l l l l l l | l }
   Method & time & jacket & dress & skirt & top & trousers & purse & shoe & mAP \\
   \hline Faster R-CNN end2end & 141ms & 83\% & 73\% & 75\% & 69\% & 84\% & 63\% & 85\% & 76.3\% \\
   SSD 300x300 & \textbf{30ms} & 83\% & 73\% & 75\% & 76\% & 85\% & 66\% & 86\% & 77.7\% \\
   SSD 500x500 & 70ms & 84\% & 74\% & 76\% & 76\% & 86\% & 68\% & 87\% & 78.9\% \\
   R-FCN ohem ResNet-50 & 96ms & 88\% & 78\% & 80\% & 77\% & 90\% & 68\% & 90\% & 81.6\% \\
   R-FCN ohem ResNet-101 & 137ms & 89\% & 78\% & 81\% & 78\% & 91\% & 69\% & 90\% & \textbf{82.3\%} \\
    \\
   \end{tabular}
  \end{center}
  \caption{The average precision calculated on each object class and the mean of the APs.
   The bounding boxes are considered only if the intersection-of-union (IOU) is larger than 50\%. The execution times are measured on an NVIDIA Quadro M6000 GPU with 24GB memory.}
   \label{table:timings}
  \end{table*}
  
Removing the fully connected section from Faster RCNN makes the training unstable. Hence, to guaranteee stability the ROIs are subdivided in a 7x7 grid and 49 losses are calculated for each category. Such a scheme allows each part of the grid to recognize features that are typical of that part of the object. Empirically, whilst  one does not gain much on classification accuracy by training the network without the fully connected layer, this allows us to use the OHEM method.

\begin{figure}[!t]
 \centering
 \begin{subfigure}[b]{0.4 \textwidth}
  \includegraphics[width=\textwidth]{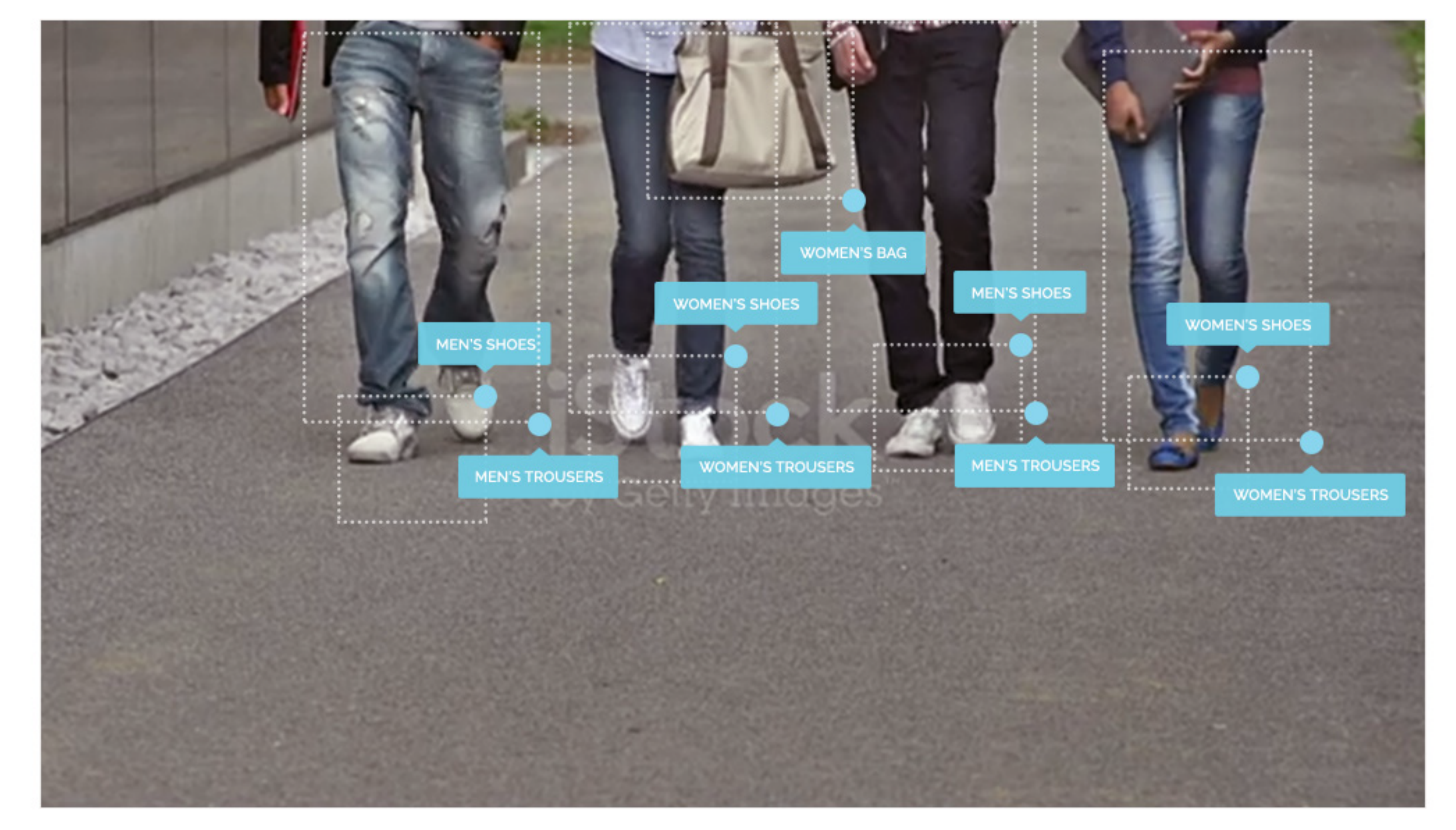}
  \label{fig:localisation_clothing}
 \end{subfigure}
 \begin{subfigure}[b]{0.4\textwidth}
  \includegraphics[width=\textwidth]{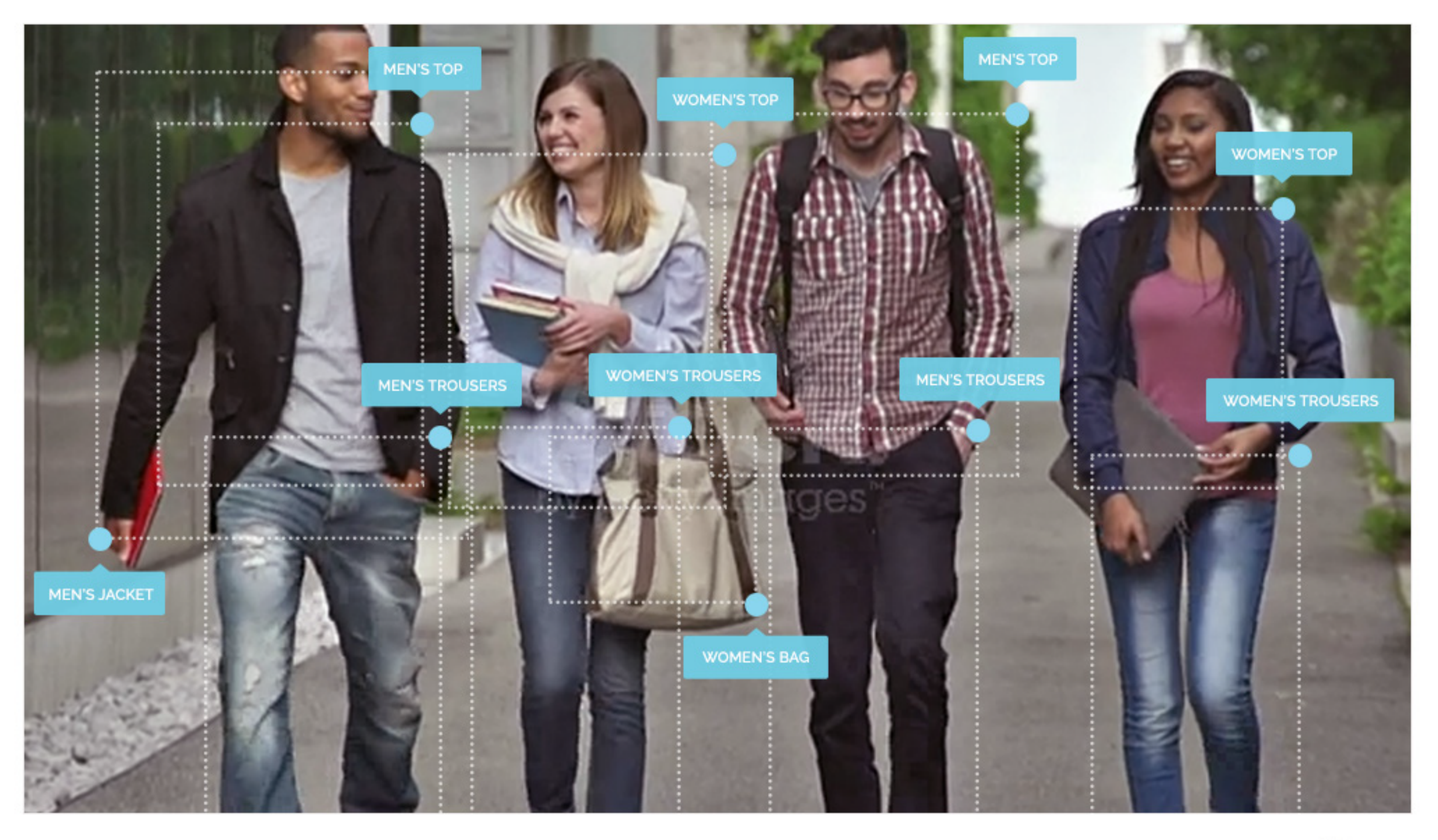}
  \label{fig:localisation_clothing}
 \end{subfigure}
 \caption{Examples of localisation on snapshots of a video. (top) This shows the items that are detected from the top of the frame. (bottom) This shows the items that are detected from the bottom of the frame.}
 \label{fig:localisation}
\end{figure}

To evaluate the performance of the three networks we  use a dataset containing 45,000 Street Style images (Figure \ref{fig:localisation}; Table \ref{table:timings}) where seven fashion categories have been manually annotated with a bounding box (bb) around the object. The categories used are: jackets (18k bb), dress (9k bb), skirt (15k bb), tops (30k bb), trousers (13k bb), handbags (23k bb) and shoes (50k bb). The 45k images are split into two sets where 40k images are used for training and 5k images are used as a test set.

To train the networks we used the default parameters chosen in the original papers.
Table \ref{table:timings} shows the average precisions (APs) of the results of the models.
Figure \ref{fig:localisation} shows the object detection of the R-FCN ResNet101 model on one image of the validation set. The two newer methods (SSD and R-FCN) both improve over Faster R-CNN. SSD is especially suited when speed is the main concern. The smaller (300x300) achieves realtime performance. The bigger R-FCN is slower, but achieves a higher precision (Table \ref{table:timings}). \\

\subsection{Architecture/Implementation}
\label{subsec:localisation_architecture}

The localisation service was designed to be \textit{horizontally scalable} in order to handle a high number of requests per second. Such scalability was made possible by employing a load-balanced micro-service architecture. The micro-service architecture provides a method of developing software applications as a suite of small, modular and independently deployable services in which each service runs a unique process and communicates through a lightweight and well-defined mechanism. 

The localisation micro-service is written in C++14 and uses POCO libraries (https://pocoproject.org) to implement a RESTful (Representational State Transfer) API. The service accepts a HTTP multipart request containing a query image and a JSON (JavaScript Object Notation) specifying various parameters. The image is then processed via a pre-specified deep-learning architecture (for example, a R-FCN network), and the localisation results are returned to a client in a JSON containing the classifications, confidences and coordinates of localised items.

A load-balancer is used to distribute the requests among any number of micro-service instances distributed across any number of GPU servers. Currently, we use servers with 4x NVIDIA GeForce GTX 1080, each featuring 2560 cores and 8GB of memory. Each server can host up to 16 instances (4 per GPU) while each instance can process approximately 5 requests/second resulting in a total of 80 requests/second per server. Additional instances can be dynamically deployed on more servers to cope with increasing throughput requirements. The deployment of new instances is straight forward since each micro-service runs from within a Docker container that guarantees that the software will always run the same, regardless of its environment. This is because each Docker container comes with the source-code, runtime executable, system tools and libraries that are needed to run the localiser micro-service.

At a very high-level, all queries are sent to a single load-balanced endpoint. The load-balancer then distributes these requests between a number of docker containers that is further distributed amongst a number of GPU-enabled servers. Each docker container runs a software load-balancing (\textit{haproxy}), which further distributes the requests amongst the number of localiser instances running within a single docker. The maximum number of these instances depend on various constraints, such as GPU compute and memory capacity as well as the size of the model used for localisation. Different docker containers can run on the same or different GPUs as well as run different models for localisation. Various localisation services are made accessible via dedicated ports.

\section{Image retrieval approach}
\label{sec:retrieval}

After having localised a specific object (e.g., coat, jacket, etc.) corresponding to one of the seven categories used for multi-product search, image retrieval is performed against a particular database of inventory items. This retrieval is performed in terms of \textit{visual similarity}. The main factors that make clothes perceptually similar are their colour and texture properties. Such properties need to be extracted in a way that is relevant to human perception. In this context, simple approaches as colour histograms are not adequate. Furthermore, relevant features must be extracted in a compute and memory efficient manner. 

Building upon prior work on psycho-physics and human neuroscience, our algorithm encodes the texture and the colour of the image in a similar way as the human brain does. This information is encoded and the matching against a large-scale dataset is performed in an efficient way. Two  steps are required to perform retrieval. First of all we extract a signature for every image. Second, we match the signature of a query image to a large dataset of images.

\subsection{Architecture for signature extraction}
\label{subsec:retreival_colour_descriptor}

In order to build a signature of an image, the following steps are required: detection of points of interest (key-points) in the image, feature extraction and encoding.
The key-point detection aims at identifying the most important locations on the image. Features are extracted from the patches around the key-points, which capture the texture and colour characteristics of the patch. In order to have an efficient protocol to match the descriptors of an image against a large database of images, feature encoding is required. A bag of words model \cite{Fei-Fei2005} is used to convert the descriptors of all patches  to codeword vectors. This gives a unique signature for each image, which is further bit-encoded to save memory space.

A point of interest in an image is a distinctive location in an image that can be robustly localised from a range of viewpoints, rotations, scales, and illuminations. In order to find these key-points in every image we use biologically-inspired non-linear orientation channels, as described in \cite{Bharath2014}. The key-points are extracted using a pyramid structure, i.e. four scales are used for each image. These scales are used in order to capture information on different level of abstraction -- a harmonic (multi-scale) representation for each image is constructed. For each key-point we also construct an associated saliency; this indicates how distinct that point is. A threshold is defined to select the most salient key-points, and a maximum of 512 keypoints is used.

For every key-point, a patch of a fixed size of 32x32 on each scale is selected; features are then calculated for each patch. The algorithm  captures texture and colour information for every patch by applying specific filters to the image data. This enables it to identify texture characteristics and colour information. The processing after the extraction of the texture information includes weighting the texture data of each one of the colour channels with intensity data of the image patch.  Texture and colour information is combined to obtain a descriptor which allows similar image patches to be identified.

Based on human visual perception, a set of steerable complex wavelet filters \cite{Bharath2014} are used to extract texture features such as edges and lines/bars in different scales and orientations. If no features are identified, for example, if the image is monochromatic with each pixel intensity being the same, the image will still have a texture, albeit a smooth and uniform one. To capture the colour information, the CIE Lab \cite{Hunter1948} colour space is used; this mimics the human brain's perception of colour. Particularly, each colour patch is first converted into CIE Lab colour space and normalized to a range of $[0,1]$. 32 band pass filters are applied for every colour channels generating a total of 96 filter responses. Texture weighted colour histograms are  subsequently generated. In the end, we obtain a 576 dimensional colour-texture descriptor (3 colour channels x 6 bins x 4 scales x 4 directions x 2 values).

\begin{figure}[!t]
 \centering
 \begin{subfigure}[b]{0.2\textwidth}
  \includegraphics[width=\textwidth]{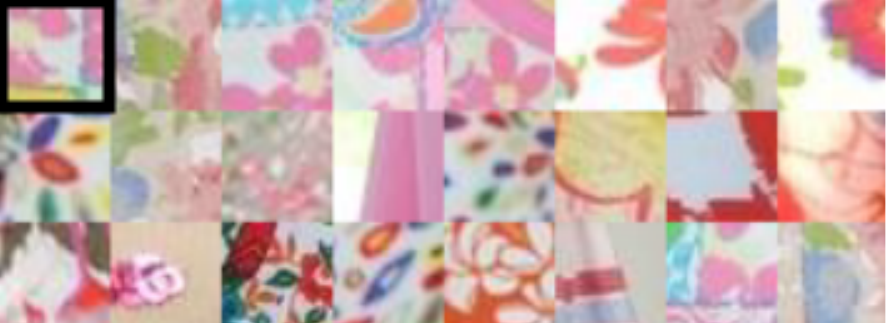}
  \label{fig:ColHist12}
 \end{subfigure}
 \begin{subfigure}[b]{0.2\textwidth}
  \includegraphics[width=\textwidth]{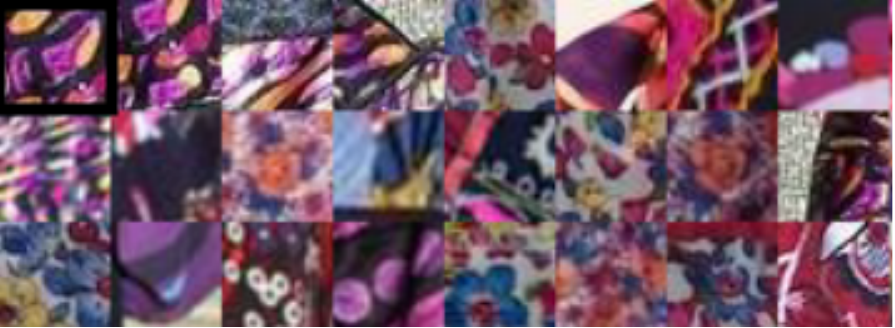}
  \label{fig:ColHist4}
 \end{subfigure}
 \begin{subfigure}[b]{0.2\textwidth}
  \includegraphics[width=\textwidth]{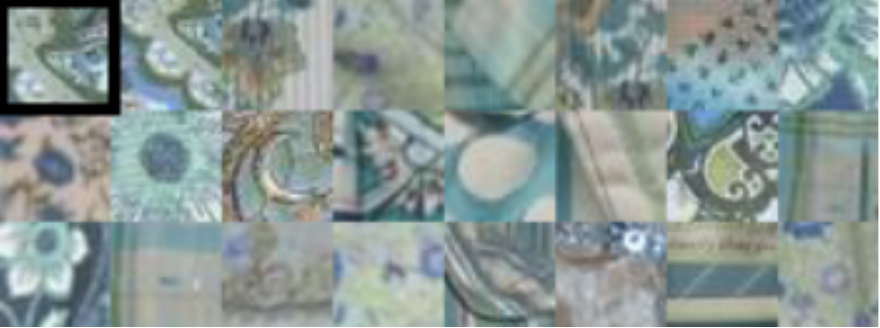}
  \label{fig:ColHist5}
 \end{subfigure}
 \begin{subfigure}[b]{0.2\textwidth}
  \includegraphics[width=\textwidth]{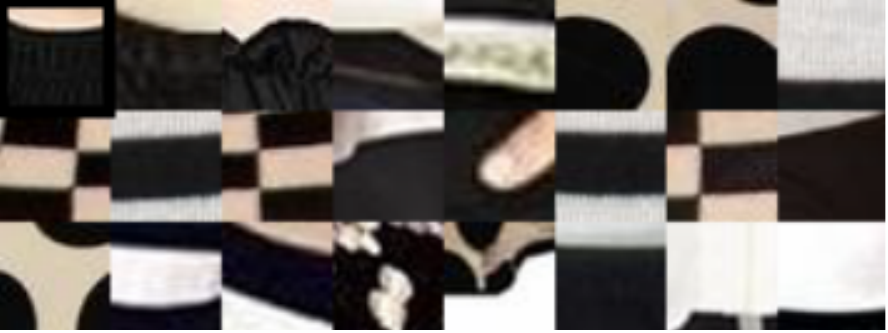}
  \label{fig:ColHist14}
 \end{subfigure}
 \begin{subfigure}[b]{0.2\textwidth}
  \includegraphics[width=\textwidth]{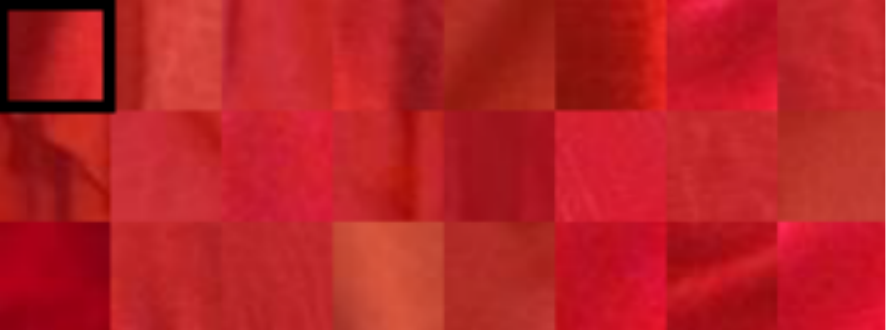}
  \label{fig:ColHist10}
 \end{subfigure}
 \begin{subfigure}[b]{0.2\textwidth}
  \includegraphics[width=\textwidth]{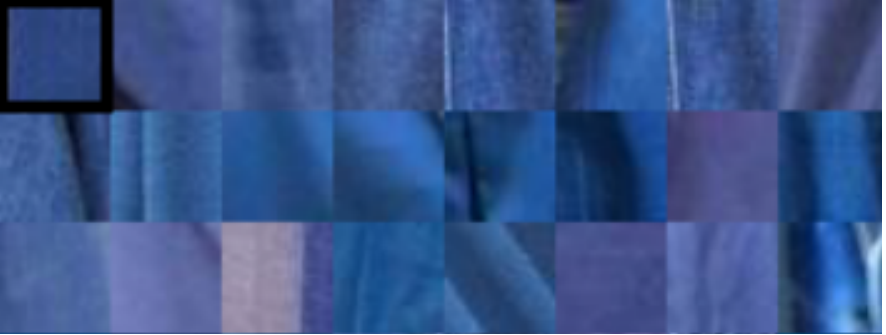}
  \label{fig:ColHist11}
 \end{subfigure}
 \begin{subfigure}[b]{0.2\textwidth}
  \includegraphics[width=\textwidth]{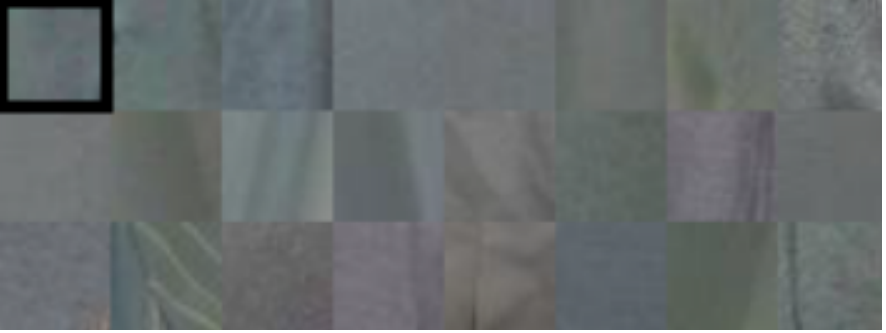}
  \label{fig:ColHist8}
 \end{subfigure}
 \begin{subfigure}[b]{0.2\textwidth}
  \includegraphics[width=\textwidth]{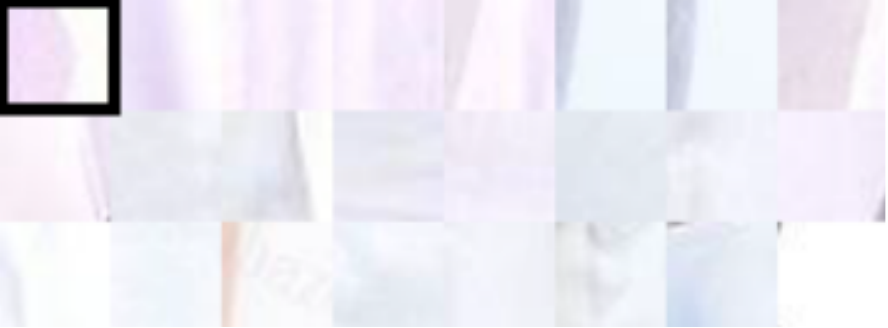}
  \label{fig:ColHist9}
 \end{subfigure}
 \caption{Example results based on the colour-texture descriptor on a patch level. The first patch, with the black outline, on the top left of the requisite image is the query, and the results are arranged in order of similarity. The first four examples show some example where both the texture and the colour of the patch are taken into account. The last four examples show that the algorithm is able to capture  subtle texture differences as well as colour variations.}
 \label{ColHist}
\end{figure}

In particular, the texture intensity data describes not only the relative difference between different points in an image patch, but also the absolute intensity of the texture. This enables a descriptor for  finding similar images, for example as that perceived by a human viewer, in terms of both texture, intensity and colour. For illustration we present an example scenario in Figure \ref{ColHist}. The key-point and feature extraction is implemented using CUDA for compute time efficiency -- the entire calculation takes on an average 220 ms.

In order to encode the feature vectors from the different patches in one signature, a codebook needs to be defined. This is formed from a large number of descriptors of various type of images. The centres of the clusters, which emerge from a $k$-means algorithm, are used to map the vector values. The feature vectors of each image are mapped to a signature using the codebook, which consists of 5000 centres. This results in a signature that is approximately 3.5kB after being bit-encoded. In order to compare the similarity between two images the ${\chi}^2$ distance between the signatures is calculated. The next section discusses how the matching is operationalized in an efficient manner.

\subsection{Architecture for matching}
\label{subsec:retreival_architecture}

After the bag of words (BOWs) have been produced as described in Section \ref{subsec:retreival_colour_descriptor}, they are stored in a specific data structure that maximises search and retrieval times. In particular, all of the BOW files created with the customers' images are processed and an ``Inverted Index" is produced. Here, we explain how this index is created and utilized at query time by a distributed in-memory data-grid.

To understand the structure of the ``Inverted Index" we need to first understand the data that needs to be stored. In a single bag of words, for instance for the Colour-Texture descriptor, we usually have 2.5k words in a dictionary of around 5k. This means that if we were to store the BOWs in a big matrix, where on one axis we had the IDs of the images and on the other we had all the words, the matrix would have at least half of its entries being zeroes -- a sparse matrix. The logical solution is to store the data not including the zeroes and use an extra array to ``index" it. We choose to use an inverted indexing strategy, as is common practice in storing sparse matrices.

More precisely this means that the ``Index" array will point to the beginning of each ``word" in the data array. The data array is, in the most simple case, just a list of image-IDs, stored continuously for each word. By pointing at the inverted index, we can quickly find images that contain a specific word. Once we have a data array that is indexed by word, it is easy to distribute it across many machines, each of which carry a range of words. In this way, if we have enough instances, using an inverted index gives us the added benefit of making our system robust: in case one of the machines is faulty, the search would still scan through all the customers' images, and it would ignore just a small range of words present on the faulty server, thereby minimally affecting the query results. The size and distribution of the grid is fully customisable depending on the configuration and number of images one needs to store. Each  query runs in parallel in all the grid instances, each of which uses multiple threads to access the index. 

%\subsection{Results obtained}
%\label{subsec:retreival_results}
%Timing and stuff .

\section{Combining localisation and retrieval for multi-product search}
\label{sec:Combining}

In order to achieve a multi-product search based on one image or a video stream, the two methods described above are combined. The multi-product search runs sequentially -- localisation runs first based on a deep learning architecture and subsequently retrieval based on colour-texture descriptors is initiated. 

The input image gets processed by the micro-service for localisation. The output consists of the identified categories and the associated bounding boxes. There is some extra logic in place after the localisation takes place, which ensures that mutually exclusive items do not appear together. For example, if a dress, a top and a skirt are detected in the same area then the one(s) with the higher confidence are kept. This ensures that there are less false positives, and the results are more visually agreeable to the user.

\begin{figure}[!t]
 \centering
 \begin{subfigure}[b]{0.35\textwidth}
  \includegraphics[width=\textwidth]{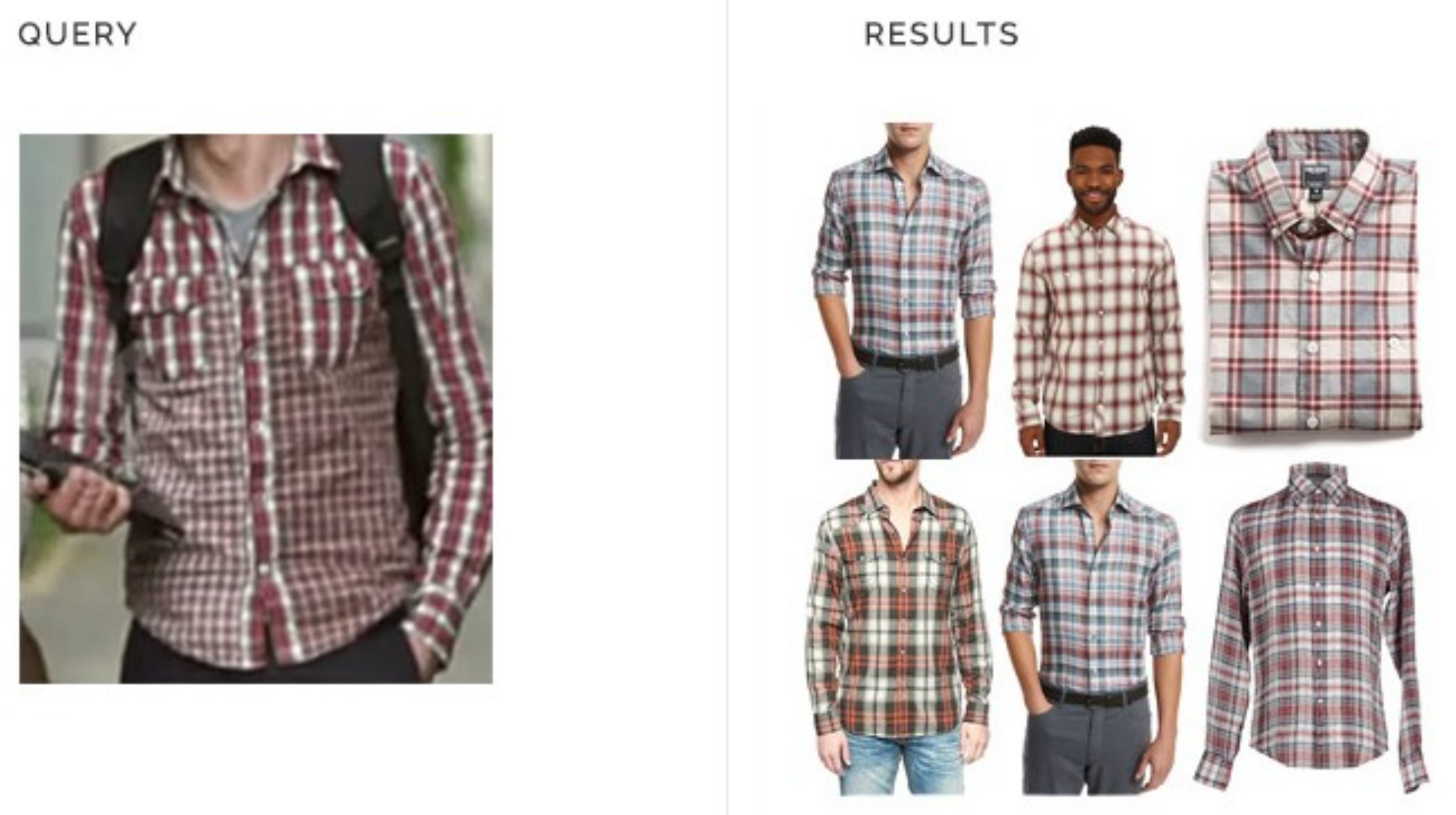}
  \label{fig:localisation_clothing}
 \end{subfigure}
 \begin{subfigure}[b]{0.35\textwidth}
  \includegraphics[width=\textwidth]{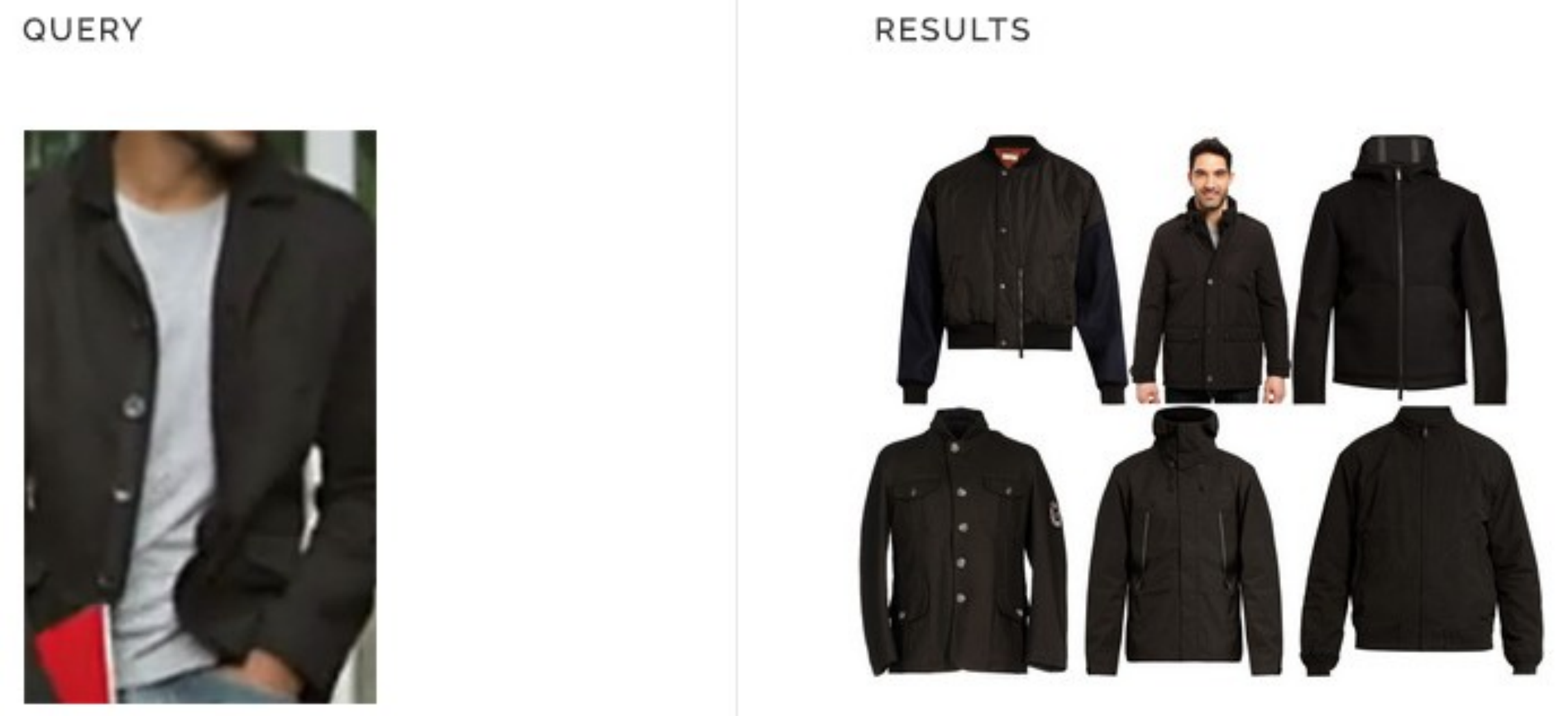}
  \label{fig:localisation_clothing}
 \end{subfigure}
 \caption{Example results of retrieval on a category of tops and a category of jackets based on the output of localisation on the snapshot of the video shown in Figure \ref{fig:localisation}. The image gets cropped based on the output of the localisation algorithm; the database is then queried on the detected category.}
 \label{Localisation_Retrieval}
\end{figure}

Each cropped image around every detected bounding box is used as a query against the identified category. The final retrieval results for every category are returned. Using the same snapshot of the video in Figure \ref{fig:localisation}, the category of each item localized (query object) is returned followed by items that are similar to this query object. Similar items obtained from two queries are shown in Figure \ref{Localisation_Retrieval}. 

For each potential category that can be detected by using the trained deep learning model, there is an associated database. Splitting the database according to the object type results in more accurate inference. On the other hand, performing retrieval in a very large database that includes all object types items deteriorates the performance. We use in this case 7 separate databases, one for each product, against which similar items can be found. This means that there are 7 different database IDs and associated inverted files. In our experiment each database includes 100 thousand to 2 million images. Certain categories, like tops, have a vast amount of data available, while for others, like purses, data is limited; this results in databases with varying sizes. The queries against the different databases are performed in parallel. On one hand this keeps the overall timing below 1 $s$ on the other hand this timing is not affected by the number of items detected. In addition, the timing of each  query against each database is also approximately similar regardless of its size, since the retrieval is performed under a distributed architecture. Overall the process achieves high accuracy and  compute time efficiency without requiring any user intervention.

\section{Conclusions}
\label{sec:conclusions}

In this paper we have described \textit{`Cortexica's multi-product search'}, a framework for large-scale localisation, classification and retrieval. This is being used for a wide variety of  commercial applications -- from fashion industry, health and safety critical applications to medical imaging. It harnesses software scalability using Docker containers whilst hardware scalability is achieved using GPUs that are deployed on a wide variety of cloud computing providers. Due to the hot-swappable nature of our implementation, one can not only use any number of compute efficient deep-learning framework but also non-parametric Bayesian classification and regression algorithms in near future. This gives us immense flexibility to choose the most efficient statistical and compute efficient algorithm for the application of interest. Our current research aims to have a similar architecture not just for image or video data-streams but also add an added layer of security for sensitive data by employing privacy-preserving machine learning techniques.

\bibliography{large_TMM}
\bibliographystyle{plain}

\end{document}